# Requirement Formalisation using Natural Language Processing and Machine Learning: A Systematic Review

Shekoufeh Kolahdouz-Rahimi[1] [a], kevin Lano[2] [b] and Chenghua Lin[3] [c]
[1]School of Computing, University of Roehampton, London, UK
[2] Department of Informatics, King's College London, London, UK
[3] Department of Computer Science, University of Sheffield, UK
shekoufeh.rahimi@roehampton.ac.uk, kevin.lano@kcl.ac.uk, c.lin@sheffield.ac.uk

Abstract:
Improvement of software development methodologies attracts developers to automatic Requirement Formalisation (RF) in the Requirement Engineering (RE) field. The potential advantages by applying Natural Language Processing (NLP) and Machine Learning (ML) in reducing the ambiguity and incompleteness of requirement written in natural languages is reported in different studies. The goal of this paper is to survey and classify existing work on NLP and ML for RF, identifying challenges in this domain and providing promising future research directions. To achieve this, we conducted a systematic literature review to outline the current state-of-the-art of NLP and ML techniques in RF by selecting 257 papers from common used libraries. The search result is filtered by defining inclusion and exclusion criteria and 47 relevant studies between 2012 and 2022 are selected. We found that heuristic NLP approaches are the most common NLP techniques used for automatic RF, primary operating on structured and semi-structured data. This study also revealed that Deep Learning (DL) technique are not widely used, instead classical ML techniques are predominant in the surveyed studies. More importantly, we identified the difficulty of comparing the performance of different approaches due to the lack of standard benchmark cases for RF.

## 1 INTRODUCTION

Productive management of Requirement Engineering (RE) accelerates the process of software development. Requirement Formalisation (RF) relates to the process of transforming requirement in natural language to specific formal notations by removing ambiguities. Formal specification of requirement is applicable in different stages of software development specially in the validation phase. Manual formalisation of natural language requirements is an error-prone and time consuming task and infeasible for complex systems (Zaki-Ismail et al., 2021). To this end many automatic and semi-automatic approaches have been introduced to formalise requirement by applying Natural Language processing (NLP) techniques (Rolland and Proix, 1992; Ryan, 1993). Additionally, leveraging Machine Learning (ML) and Deep Learning (DL) techniques in this domain push the research field forward.

In the last decade, there has been a noticeable increase in the number of papers using NLP and ML techniques for RF. Each research applied particular technique and mostly there is no comprehensive guidelines for the reason of applying those techniques. A large number of research reviews have been carried out to survey this domain (Alzayed and Al-Hunaiyyan, 2021). To overcome the limitations of existing studies, in this paper we conducted a systematic mapping study of NLP and ML approaches for RF considering the guidelines presented by Kitchenham and Charters (Brereton et al., 2007), (Keele et al., 2007), and Petersen et al. (Petersen et al., 2015). We investigated 47 studies from an initial set of 257. The papers are selected from commonly used libraries including ACM Digital Library, IEEE Xplore, ScienceDirect, Springer Link, and Scopus. The search result is filtered by defining inclusion and exclusion cri-teria to decide whether a publication found in the search should be included in the study or excluded. Three research questions were formulated for our Systematic Literature Review (SLR). By answering these research questions, the state of the art of RF is evaluated. We identify current challenges of community and provide guidelines to address those challenges. The actions that were taken by the author of this paper as part of MDENet project [1] are also discussed. Finally, for further maturity of the research field, future research directions in this area are provided.

[a] https://orcid.org/0000-0002-0566-5429
[b] https://orcid.org/0000-0002-9706-1410
[c] https://orcid.org/0000-0003-3454-2468

The remainder of this paper is organised as follows: In Section 2, related review papers in RF using NLP and ML are summarised. Section 3, presents background information related to this domain. The applied research method in this study is described in Section 4. Section 5 outlines the key findings of our study. Discussion on action taken by the authors to address the challenges is provided in Section 6. Finally, conclusion and future direction of research are provided in Sections 7 and 8.

## 2 Related Work

A comprehensive survey in application of NLP to RE is provided in (Zhao et al., 2020), by considering 404 works. This paper emphasise is on the insufficient application of NLP techniques for RE studies in industrial cases. Importantly, lack of expertise in selecting appropriate NLP techniques in RE domain is discussed. Although challenges are introduced in this work, practical solutions are not provided. Additionally, investigating ML techniques in this domain is not the main focus of research.

A survey in application of NLP techniques to requirements in the form of user stories is provided in (Raharjana et al., 2021) and potential advantages of those techniques in RF domain is discussed. A comprehensive classification based on the uses of NLP techniques for user stories is provided and a model is recognised as a common target for RF in the form of user stories. However, ML techniques for RF are not considered in this research and in general the main challenges of the domain are not sufficiently discussed. Additionally, input datasets for user stories in investigated papers are not provided.

Yalla and Sharma (Yalla and Sharma, 2015) survey the current literature that leverages RE and NLP for different phases of software development. However, only limited future research direction and guidelines are provided in this work. Selected articles that generates UML diagram by applying NLP techniques are investigated in (Dawood et al., 2017; Abdelnabi et al., 2021). These works emphasize on immaturity of research area as most of the current processes are not automated. Advantages and disadvantages of different studies that generates UML diagram by applying heuristic rules are identified in (Ahmed et al., 2022). This work highlighted the noticeable application of ML in this domain.

The are many related literature studies in the RF domain. However the the challenges of the domain are not deeply recognized and guidelines for addressing those challenges and clear research direction are limited in most studies. This proves the immaturity of the research area and its potential for further improvement. Therefore, the main aim of this research is to identify current challenges of community by classifying applied techniques and provide practical guidelines for addressing those issues.

## 3 BACKGROUND

In this section, the related concepts to this research including RE, NLP and DL are explained.

### 3.1 Requirement Engineering

RE is an important process in software development for discovering stakeholders needs and classifying them for other phases of software development (Pohl, 2010). Different challenges have been identified in the RE phase. One of the unavoidable challenges in this domain is ambiguity in natural language. This issue is the main focus of many studies and various approaches have been introduced to solve this problem. NLP is one of the widely used technique that supports RE in this dimension (Tukur et al., 2021; Zhao et al., 2020).

#### 3.1.1 Requirement Formalisation

Formalising requirement is one of the key tasks in RE that is automated by applying different NLP and ML techniques and tools. It enables to translate the requirement in natural language into a structured formal form (e.g., UML modeling diagrams). This transformation reduces the ambiguities of natural language and provides convenient way for validation and verification (Scho¨n et al., 2017; Tukur et al., 2021).

[1] https://mde-network.com/

## 3.2 Natural Language Processing

NLP is an area of research in Artificial Intelligence (AI) that enables computer to process a large amount of structured/unstructured data in natural language that exist in today's world. Different NLP method, approach, process and procedure are introduced to process data in different phases of RE. Tokenization, POS tagging and dependency parsing are the commonly used techniques in this domain (Kulkarni and Shivananda, 2019).

## 3.3 Machine Learning

ML is one of the core technical terms in Artificial Intelligence (AI), which refers to the learning and identification of patterns from examples and existing data (Samuel, 1967). Different algorithms and processing techniques are introduced in this domain. ML is organised in three main types including Supervised learning, Unsupervised learning and Reinforcement learning (Cepukenas et al., 2015; Li et al., 2020). ML is used to solve various real world issues. Applying ML techniques improves different RE tasks such as classification and model extraction. A real word example of application of ML is for RE.

DL is a ML technique, which is based on Artificial Neural Network (ANN) (Goodfellow et al., 2016) and is applicable in variety of domains including RF.

# 4 RESEARCH METHOD

The provided guidelines by Kitchenham and Charters (Brereton et al., 2007), (Keele et al., 2007), and Petersen et al. (Petersen et al., 2015) are applied for systematic mapping study in this research. The following research questions are the main target of this research.

- **Q1:** What are the most commonly used NLP/ML approaches for automatic/semi-automatic RF?
- **Q2:** What are the input and output of RF approaches?
- **Q3:** What are the gaps and deficiencies in existing RF work?

The three phases of study protocol including planning, conducting and reporting are explained in the following sections.

Table 1: Terms for Selecting Relevant Research Studies

| Group | Term |
|---|---|
| A | Natural Language Processing<br>Natural Language, NLP |
| B | Machine Learning<br>Deep Learning |
| C | Requirement Formalisation<br>Model Generation<br>UML Generation<br>OCL Generation<br>Usecase or Use case Diagram Generation<br>Class Diagram Generation<br>Sequence Diagram Generation<br>ER Diagram Generation<br>Activity Diagram Generation |

## 4.1 Review Planning

The review process and search strategy are explained in this part. To provide comprehensive coverage of existing publication most major publishers in Software Engineering are investigated. The following research libraries are used to find the related studies for RF.

- ACM Digital Library (http://dl.acm.org)
- IEEE Xplore (http://ieeexplore.ieee.org)
- ScienceDirect (http://www.sciencedirect.com)
- Springer Link (http://www.springer.com)
- Scopus (http://www.scopus.com)

According to the objectives of this study and research questions three terms were selected in this paper. Each term includes different keywords and at least one of the keywords has to be presented in a paper. Table 1 presents the list of selected terms in this research.

To identify the largest number of studies in the domain of RF, the following search string is followed:

Search String= $(A \lor B \lor (A \land B)) \land C$

Additionally, inclusion and exclusion criteria to decide which of the selected articles should be considered as primary studies and which ones should be excluded are defined as follows:

### 4.1.1 Inclusion Criteria

- Published between January 2012 and March 2022
- Publications that generate model or any formalisation from requirement
- Publications in peer-reviewed journals, conferences, and workshops

- Publication in English

#### 4.1.2 Exclusion Criteria

- Publications not written in English
- Publications before 2012
- Summary, survey, or review publications
- Non peer-reviewed publications
- Publications not focusing on RF
- Books, web sites, technical reports, pamphlets, tutorials, duplicate papers, and white papers.

In this research abstracts, titles, and keywords of papers are evaluated according to the inclusion and exclusion criteria. Furthermore, for some cases the whole text of the paper is also investigated.

### 4.2 Review Conduction

The review conduction stage presents the selection process for LR in this research.

#### 4.2.1 Article selection:

This phase is divided into three sub tasks including, pilot study, article selection and quality assessment of selected primary studies.

**Pilot study**: A pilot study is carried out to investigate the reliability of provided selection criteria as suggested by Kitchenham and Charters (Brereton et al., 2007), (Keele et al., 2007), and Petersen et al. (Petersen et al., 2015) before selection of primary articles. In this stage five papers are selected by the first and second authors. These articles are investigated by third author, who is an expert in NLP domain and was not involved by the search process by considering the inclusion and exclusion criteria. A satisfactory result is presented from pilot study, which proves the suitability of the defined criteria in this research.

**Primary study selection**: In this part, the relevant articles are searched using the provided search string. The result of this selection is presented in Figure 1. For the initial search process 257 results are recognised. To refine the selection papers in the next iteration, titles, abstracts, keywords, introduction and conclusion sections are reviewed. As a result 115 papers are removed from the list of selection and then the rest are kept for the next iteration. Following that by applying inclusion and exclusion criteria 71 papers are rejected. Next for this iteration, the content of the paper is investigated deeply and 24 papers are rejected. Finally, 47 studies are remaining. Figure 3 presents the distribution of the resulting papers in each year. As can be seen in this figure, in 2021 the domain gained more interest and highest number of publication were published. Additionally, Figure 2 indicates the publication type of result papers. Conference are the target of publication in most studies.

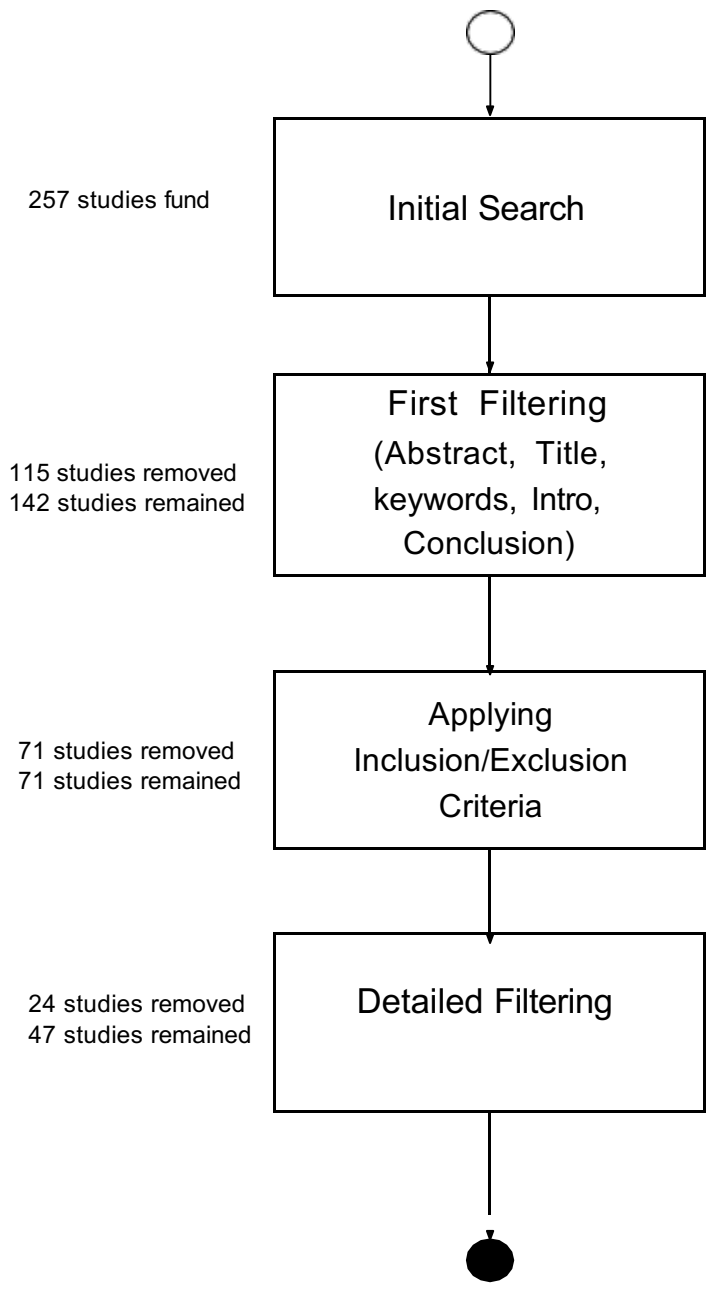

Figure 1: Primary studies selection process

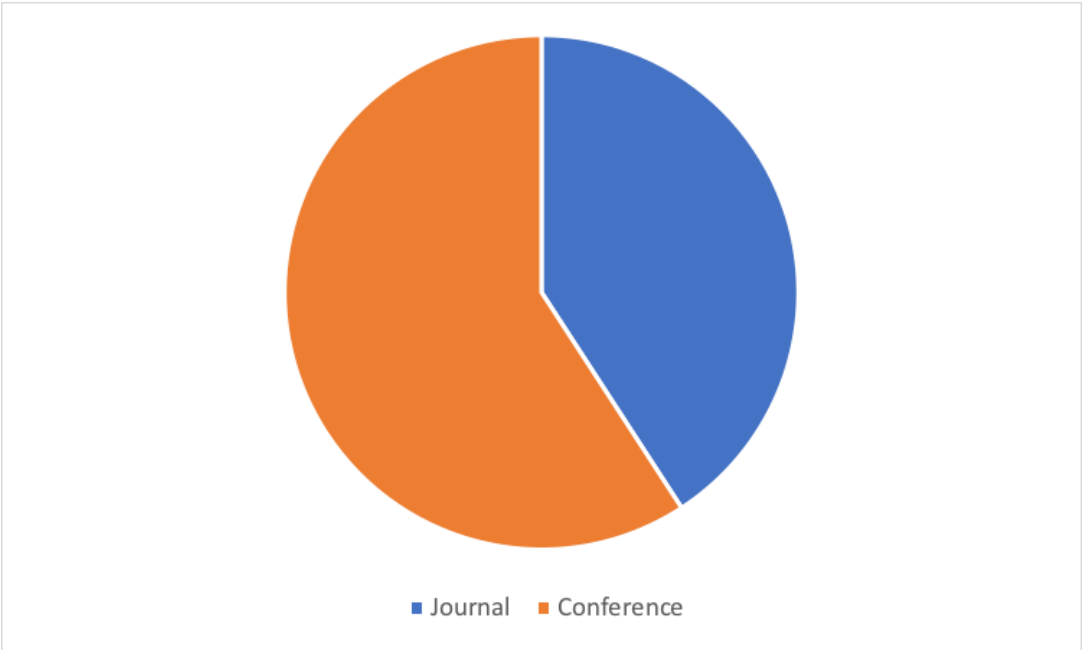

Figure 2: Primary studies per publication type

**Quality Assessment**: The quality of selected study are also assessed in this research. Therefore,

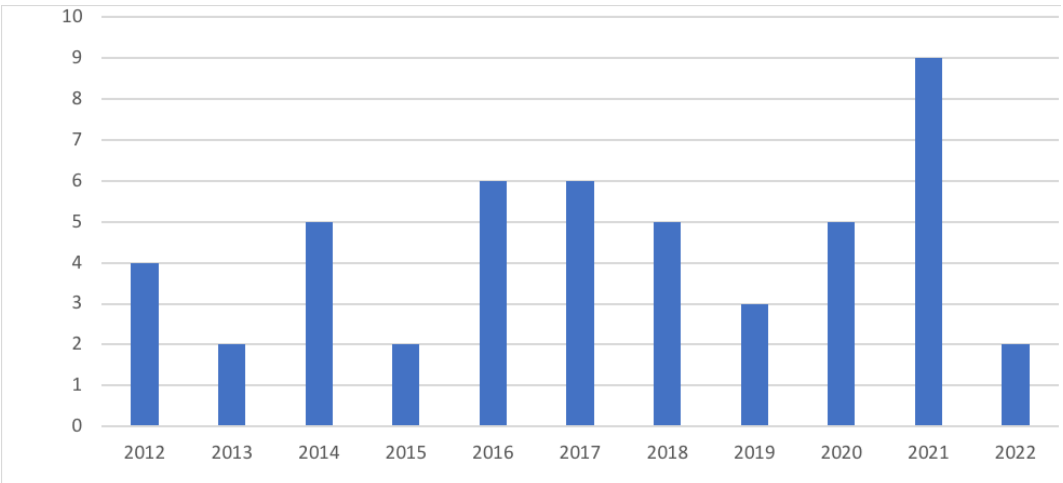

Figure 3: Distribution of NLP papers in each year

Table 2: Quality assessment Questions

| QID | Topic | Question |
|---|---|---|
| A1 | Objective | Did the study clearly define the research objectives? |
| A2 | Related work | Did the study provide a review of previous work? |
| A3 | Research methodology | Are the research methodology clearly established? |
| A4 | Validity | Did the study include a discussion on the validity and reliability of the procedure used? |
| A5 | Future work | Did the study point out potential further research? |

a checklist with four quality assessment questions are presented in Table 2. The questions are answered by the first author by selecting from 'yes', 'no' and 'partly' options.

#### 4.2.2 Data Extraction and Synthesis

To answer each research question, the data extraction process is performed by developing a predefined data extraction form in Table 3. The form enables us to record essential information of primary studies to answer each research question. The form is fill out by the first author manually and then second and third authors reviewed the results and finally the issues are fixed.

#### 4.2.3 REPORTING THE REVIEW

Based on the results of data extraction phase, the review result is presented and each research question is answered and discussed.

## 5 RESULTS

This section presents the result of review in this research. We selected 47 primary studies for SLR.

### 5.1 Summary of Studies

Selected studies for classifying NLP and ML techniques to formalise requirement are presented in Tables 4 and 5. Due to the lack of space, the rest of the studies are available in (RFr, 2022). The selected studies are summarise according to the applied NLP/ML techniques, input and output artifacts, datasets, approach, applied tools and libraries, input structure and evaluation criteria.

This section discusses the classification results of the investigated approaches according to each research question.

Table 3: Data Extraction Form

| Study data | Description | Relevant RQ |
|---|---|---|
| Title | | Study Overview |
| Author | | Study Overview |
| Year | | Study Overview |
| Article Source | | Study Overview |
| Type of Article | Journal, Conference Workshop | |
| Research goal | What is the main goal of study? | RQ1 |
| Research goal category | Model extraction/generation, Requirement formalisation, UML, Usecase, Class Activity, ER diagram extraction/generation | RQ1 |
| Research method | What research methods did the study employ? | RQ1 |
| Data | What are the datasets for evaluation of the study | RQ2 |
| Evaluation | what are the evaluation criteria in the study? | RQ1 |
| NLP techniques | What are the applied NLP techniques? | RQ1 |
| ML techniques | What are the applied ML techniques? | RQ1 |
| NLP tools | What NLP tools did the study use? | RQ1 |
| Challenges | What are the challenges of the study? | RQ3 |
| Future Work | What are the suggested future work? | RQ3 |

### 5.2 Q1: What are the most commonly used NLP/ML approaches for automatic/semi-automatic RF?

To answer this question, NLP and ML approaches applied in selected studies are deeply investigated.

#### 5.2.1 NLP techniques

Figure 4 presents the applied NLP techniques and frequency of using those techniques through out investigated papers.

- **Applied techniques**: Tokenization, POS tagging and Type dependencies are the most common used NLP techniques in those research.
- **Heuristic rules**: Majority of research applied heuristic rules for formalising requirements.
- **Frequent tools and libraries**: Stanford core NLP is the most common-used NLP tools in investigated studies.
- **Evaluation criteria**: Accuracy in terms of precision, recall and F-measure are the most com-

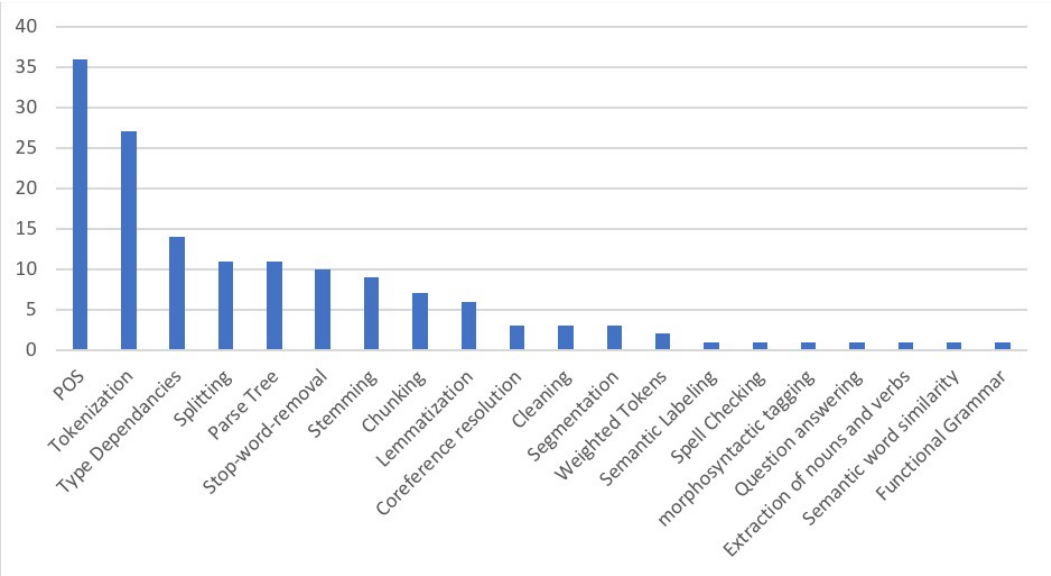

Figure 4: Frequency of NLP Technologies in Different Studies

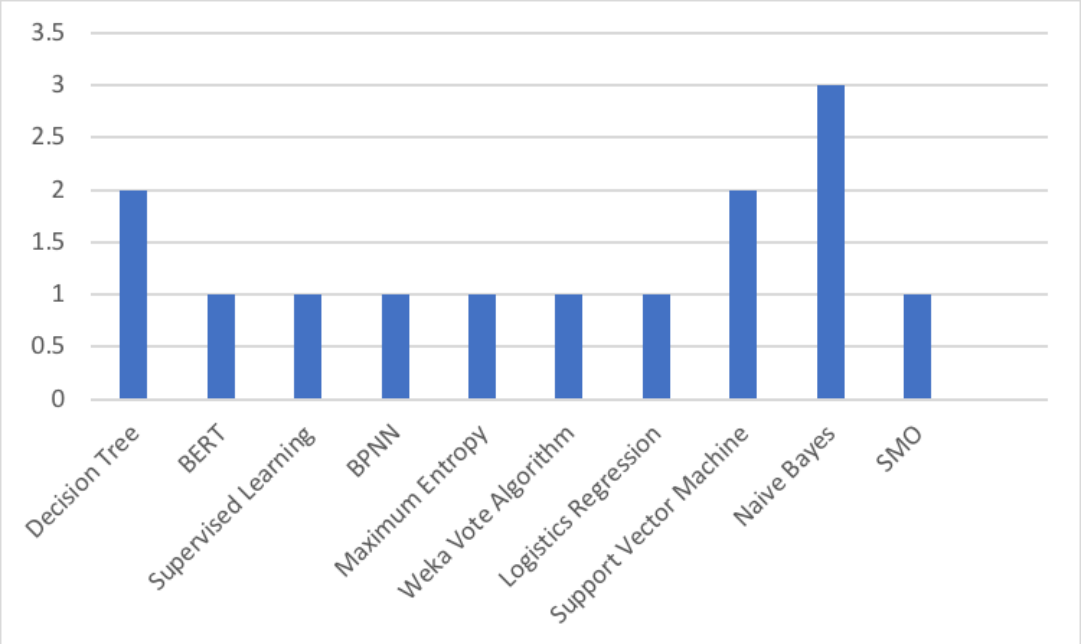

Figure 5: Frequency of DL Technologies in Different Studies

mon criteria for evaluation in selected studies.

#### 5.2.2 ML techniques

There are not many studies that apply ML techniques for RF and mostly classical ML techniques such as decision tree and Support Vector Machine (SVM) are used in those studies. The applied ML techiques and frequency of application of these techniques are presented in Figure 5. Around 20% of selected studies in this research applied both NLP and DL techniques.

### 5.3 Q2: What are input and output of RF approaches?

This question is answered according to the applied type and structure of input elements, and generated output elements.

- **Input Types**: The frequency of input types in selected studies is presented in Figure 6. English text and user story are the most commonused type in most studies.
- **Input Structure**: The unstructured English text is the most frequent input structure for most of the studies as presented in Figure 7. The inputs in the format of user story are semistructured.

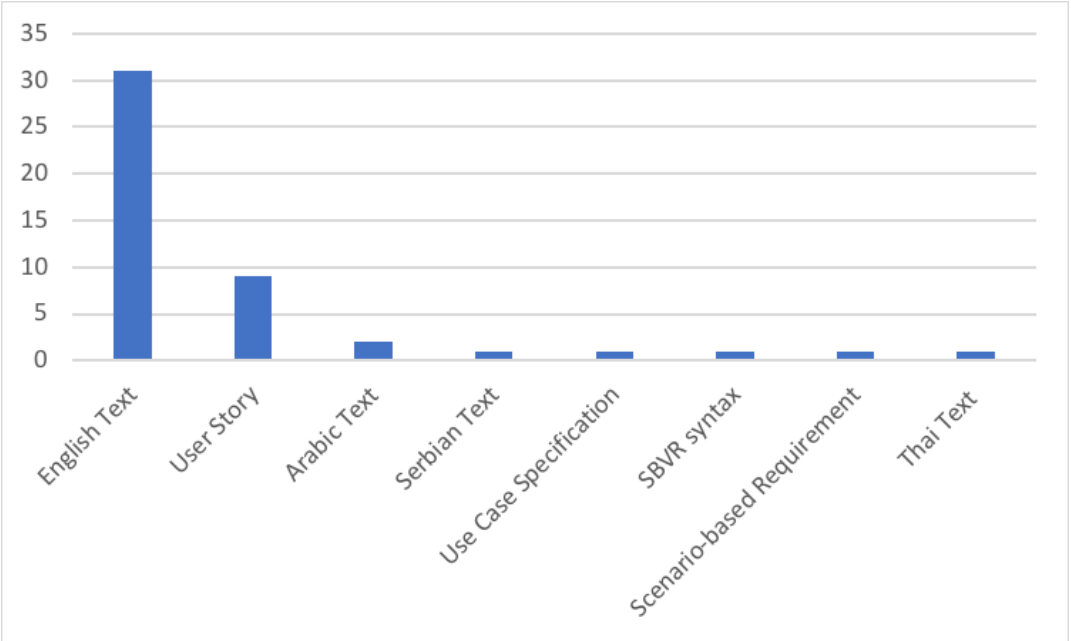

Figure 6: Frequency of Input for RF

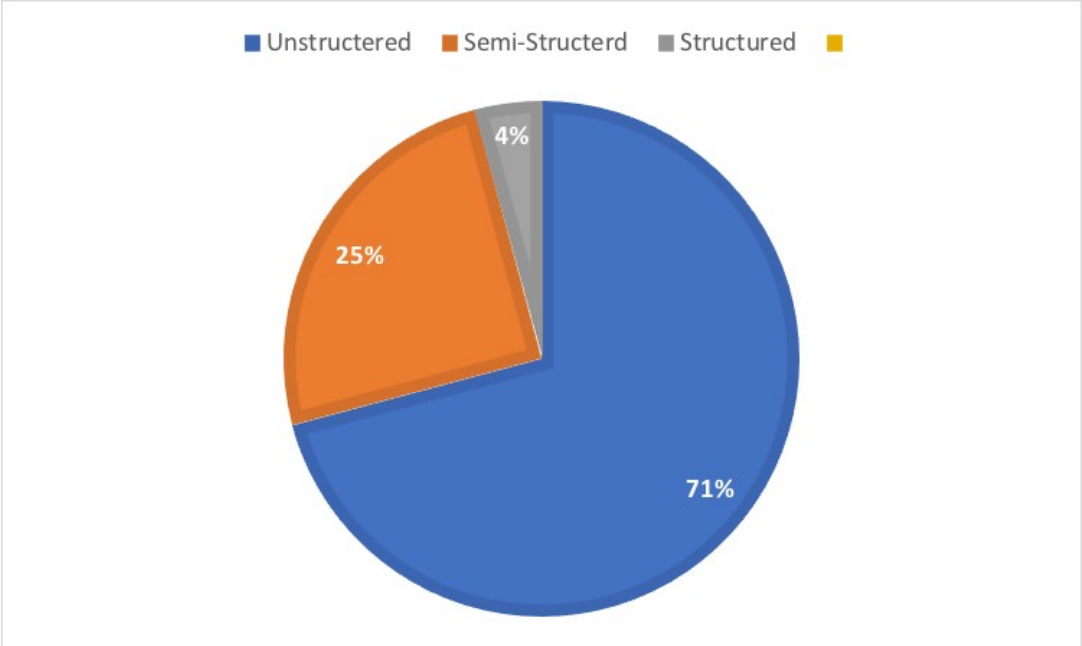

Figure 7: Structure of Input Elements

- **Output Types**: In most of studies the formalisation is in the form of UML diagrams. Figure 8 indicates that class and use case diagrams are the most common input types in these studies.

### 5.4 Q3: What are the gaps and deficiencies in existing RF work?

The RF field remains at an experimental stage, in particular evaluation of approaches is not performed in a systematic manner and it is difficult to compare different approaches. The published results of studies were often not reproducible due to unavailability of tools or data.

Investigating different works, we identified three main deficiencies for formalising requirements. A list of deficiencies are provided below:

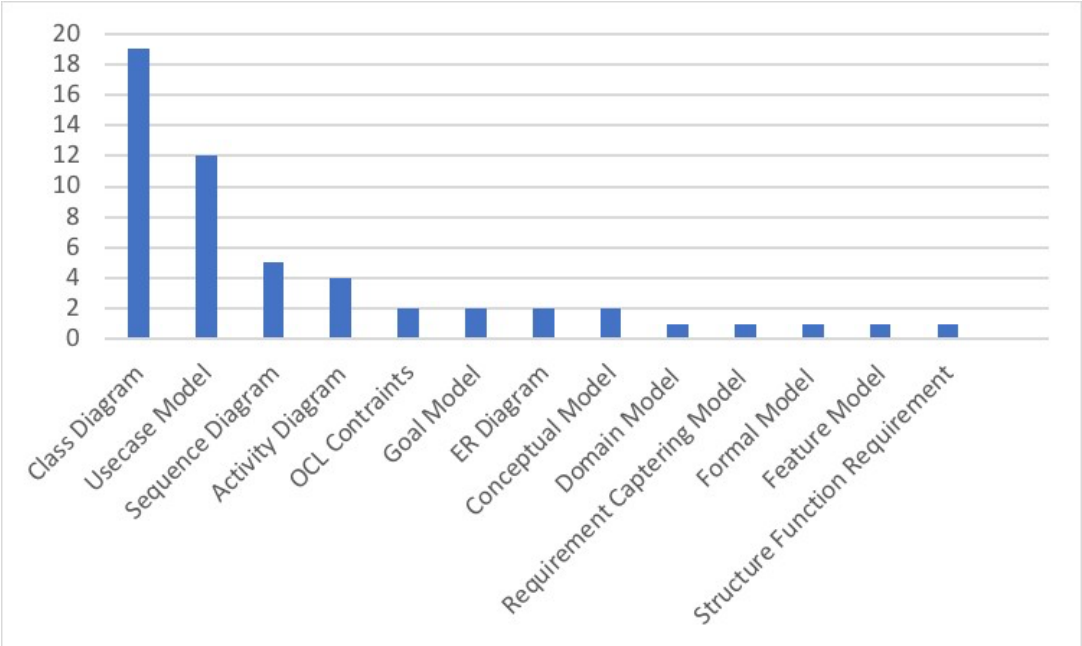

Figure 8: Frequency of Generated Output for RF

#### 5.4.1 Lack of Completeness in Heuristic rules

The performance and completeness of heuristic approaches are typically not evaluated on a broad range of input cases, thus it is not possible to determine which are the best to use in different situations. A lot of papers applied heuristic approaches and defined rules manually. It is not possible to come up with any specific number of rules for formalising requirement and generating relevant artifacts. Therefore, the provided heuristic methods in most approaches inevitably have some incompleteness and may only work well for inputs restricted to particular formats or linguistic styles. This issue is not deeply investigated in the community as there is not adequate comparative evaluation in this domain.

#### 5.4.2 Lack of application of DL

There is under-use of DL techniques, which seem to be relevant and applicable to RF tasks and could help to avoid the limitations of heuristic approaches, especially for unstructured source data. It is assumed that the limited number of training data in the community is the main reason that developers do not use DL models in different tasks. Most of the learning methods used in the community are standard methods such as decision tree or regression model. It can be concluded that most of the studies use learning in a wrong way and not exploit the full potential of deep learning techniques in this domain. Therefore, in theory application of ML provides a potential solution for different task of RF.

Consequently, there is a lot of space in application of ML and specially DL techniques to different tasks for RF to make it more mature than being experimental. Systematic comparison of DL and heuristic solutions would help to identify the appropriate techniques for particular kinds of RF tasks. Community can benefit from the DL models and tools by applying some modern learning architecture such as OpenAI Codex (Chen et al., 2021) as in these models it is not essential to have lots of labels data for performing particular task.

#### 5.4.3 Lack of Evaluation Benchmark Framework

In order to systematically compare different RF cases, standard benchmarks and evaluation criteria need to be established e.g., there are well established benchmarks in Natural Language Generation (Su et al., 2021) and Speech Processing (Yang et al., 2021). Many of the evaluation datasets cited in selected papers are no longer available. Therefore, a repository of standard cases, proposed approaches, and evaluation procedures is necessary. This is an important issue in the community and it is essential to fill this gap.

## 6 Discussion on Action Taken

Our systematic review results show open issues and research challenges for formalising a requirements. This research is part of MDENet project and some actions are taken by the authors of this paper to solve part of the issues. These actions are summerised below:

- In order to strengthen the area of RF research, we developed a DSL for NLP pipelines, based on the SQLite grammar of GitHub - antlr/grammars-v4/SQLite. This enables the high level definition of NLP pipelines for RF, independent of any particular NLP platform such as NLTK or Apache OpenNLP. Common RF processing such as POS-tagging, segmentation, chunking, parsing can be specified. A transformation from the DSL to Python was defined in order to support implementation in NLTK. Additionally, an OCL-based version of the DSL was also defined, together with supporting tools. The DSLs were evaluated by applying them to specify and implement typical NLP pipeline tasks.
- To provide a central point of reference for RF research, we established a github repository (RFr, 2022), which will hold links to state-of-the-art research in the area, evaluation cases, evaluation tools, and the results of evaluations. The repository will be a resource for the RF community and aims to improve the practical application of RF research to real-world software problems.
- In order to compare the effectiveness of RF approaches, there needs to be an established set of requirements statements that they can be applied to. We selected 25 cases of real-world requirements statements in the format of user stories, which reflect a diversity of linguistic styles and of scale, and added these to the RF repository.
- To evaluate the results of applying RF approaches to the evaluation cases, we provide tools to (i) compare the formalized models produced by an approach to manually constructed reference models for the cases, to identify a measure of similarity of these models; (ii) compare the formalized models to the source document, in order to check completeness of the formalization; (iii) to evaluate the internal quality of the formalized model. Three kinds of evaluation stratagy are presented in Figure 9. Example evaluations have been provided for three RF approaches, evaluated on two user story case studies.

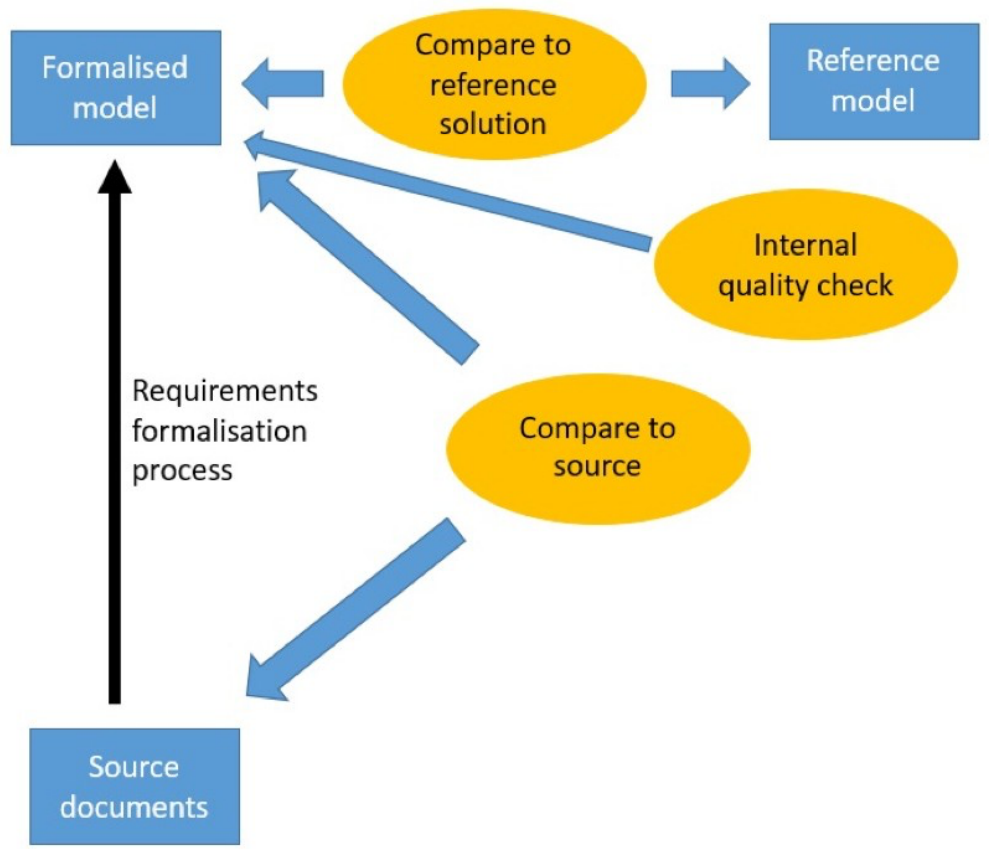

Figure 9: Three Kinds of Evaluation Strategy for RF Approaches

## 7 Future Directions

In the following we discuss directions to complete the current actions and further future work on RF.

- Generate a platform to guide the user for selecting appropriate NLP techniques occurring to their requirements. This will occur by considering more case studies and evaluating them by applying the evaluating tools specified in (RFg, ).
- We will develop the RF repository with more example case studies, including examples of unstructured requirements statements/background documentation, evaluation tools and evaluations, and publicise this in MDE forums and invite contributions from RF researchers.
- Replicability of experiments of applying NLP for RF and reproducibility of their results are currently under-addressed, and this is of particular concern for the RF community where standard benchmarks and practice do not exist. To address this issue, we will propose a shared task on the reproducibility of applying NLP in RF which aims (i) to shed light on the extent to which past evaluations are replicable and reproducible, and (ii) to propose recommendations regarding how evaluations can be designed and reported to increase replicability and reproducibility.

## 8 CONCLUSION

This research carried out a systematic survey of existing approaches for RF, including NLP and ML approaches across a wide range of applications. 250 publications were examined, and 47 specific publications were selected for deeper analysis. We identified that:

- Heuristic NLP approaches are the most common RF technique in the research, primarily operating on structured and semi-structured data.
- Deep learning techniques are not widely-used, instead classical ML techniques such as decision trees and Support Vector Machine (SVM) are used in the surveyed studies.
- There is a lack of standard benchmark cases for RF and therefore it is difficult to compare the performance of different approaches.

Table 4: Evaluation of Different NLP Approaches

| Research | NLP | Input | Output | Dataset | Approach | Tool/Library | Input Structure | Evaluation Criteria |
|---|---|---|---|---|---|---|---|---|
| (Burguen˜o et al., 2021) | Tokenization<br>Splitting<br>Stop-word-removal<br>POS | User story | Domain model | Water supply and sewage | Heuristic | GloVe<br>WordNet | Unstructured | Recall<br>Precision<br>Performance |
| (Elallaoui et al., 2018) | POS | User story | Usecase model | WebCompany (Garm, 2016) | Heuristic | Visual Paradigm<br>TreeTagger parse | Semi-<br>Structured | Applicability<br>Precision<br>Recall |
| (Zaki-Ismail et al., 2022) | POS<br>Typed-dependency<br>Parse tree | English text | Requirement capturing model (RCM) | 162 real requirements (WTT, ) | Heuristic | StanfordNLP<br>WordNet | Unstructured | Precision<br>, Recall<br>F-measure |
| (BAJWA, 2014) | Splitting<br>Tokenatioz<br>Lemmatization<br>POS<br>Parse tree | English text | OCL Constraints | Quantities, Units<br>WDimensions, and<br>Values (QUDV)<br>WebSphereBusinessModeler<br>Royal and Loyal case | Heuristic | Stanford POS tagger | Structured | Specification Coverage<br>Throughput Measure<br>Syntactic Correctness<br>Transformation Correctness |
| (Nasiri et al., 2021) | Tokenization<br>POS<br>Coreference resolution<br>Stemming<br>Stemming<br>Typed dependencies | User story | Class diagram | Inline course<br>management | Heuristic | Standord CoreNLP<br>WordNet<br>PyEcore API<br>PlantUML API<br>Visual Narrator | Semi-structured | Performance |
| (Gu¨nes, and Aydemir, 2020) | Cleaning<br>Tokenization<br>POS | User story | Goal model | - | Heuristic | - | Semi-structured | - |
| (M. Maatuk and A. Abdelnabi, 2021) | Tokenization<br>POS<br>Stemming<br>Lemmatization<br>Type Dependencies | English text | Usecase diagram<br>Activity diagram | Qualification Verification<br>System (Deeptimahanti and Babar, 2009) | Heuristic | Stanford CoreNLP | Unstructured | Applicability |
| (Ben Abdessalem Karaa et al., 2016) | Splitting<br>Tokenization<br>POS<br>Type Dependency | English text | Class diagram | - | Pattern matching? | ArgoUML | Unstructured | Recall<br>Precision<br>Overgeneration |
| (Elmasry et al., 2021) | Cleaning<br>Tokenizing<br>POS<br>Type Dependency | Support Vector Machine<br>Naive Bayes | English text | Class diagram | Public Requirement<br>dataset (Ferrari et al., 2017) | - | Unstructured | Precision<br>Recall<br>F-measure |

Table 5: Evaluation of Different ML Approaches

| Research | NLP | ML | Input | Output | Dataset/Case Study | Tool | Input Structure | Evaluation Criteria |
|---|---|---|---|---|---|---|---|---|
| (Lano et al., 2021) | POS | Decision-tree | User story | Class diagram<br>Usecase diagram | Mendeley user story datasets (WWW, )(WSt, ) | AgileUML<br>Stanford NLP<br>Apache OpenNLP | Semi-structured | F-measure<br>Effectiveness<br>Performance |
| (Akay and Kim, 2021) | Question answering | BERT | English text | Structure functional requirement | SQuAD, a crowd-sourced Microelectromechanical Systems | - | Unstructured | Applicability |
| (Elmasry et al., 2021) | Cleaning<br>Tokenizing<br>POS<br>Type Dependency | Support Vector Machine<br>Naive Bayes | English text | Class diagram | Public Requirement dataset (Ferrari et al., 2017) | - | Unstructured | Precision<br>Recall<br>F-measure |
| (Ahmed and Daleel, 2020) | Tokenization<br>POS tagging<br>Splitting | Support Vector Machine | English text | Usecase diagram | - | Confusion matrix | Unstructured | Precision<br>Recall<br>F-Measure<br>Success Rate |
| (Kashmira and Sumathipala, 2018) | Tokenization<br>Stop-word-removal | Supervised learning | English text | ER diagram | - | - | Unstructured | Recall<br>Precision |
| (Sedrakyan et al., 2022) | Sentence splitting<br>Tokenization<br>Stemming<br>POS tagging<br>Coreference resolution<br>Semantic word similarity<br>Parse Tree | Insufficient information | User story | UML diagrams | Ticket Sale System | NLA<br>WordNet<br>ROMA<br>Stanford Parser | Structered/<br>Unstructered | Recall<br>Precision<br>Fmeasure |
| (Al-Hroob et al., 2018) | Tokenization<br>Type dependencies<br>Splitting<br>POS | BPNN | English text | Usecase diagram | ATM System<br>Cafeteria Ordering System (COS)<br>library system | GATE<br>ANNIE<br>Stanford- CoreNLP | Unstructured | Precision<br>Recall |
| (Nguyen et al., 2015) | Coreference resolution<br>Type Dependency<br>Parse Tree<br>POS<br>Semantic word similarity<br>Functional Grammar | Na¨ıve Bayes<br>Maximum Entropy<br>Decision Tree | English text | Goal usecase | Online Publication System<br>Split Payment System<br>PROMISE dataset (Boetticher, 2007) | Stanford Parser (Klein and Manning, 2003)<br>GUITAR (Nguyen et al., 2014) | Unstructured | Precision<br>Recall |
| (Vemuri et al., 2017) | Tokenization<br>POS<br>Splitting | Na¨ıve Bayes | English text | Use case diagram | - | - | Unstructured | APrecision<br>Recall |
| (Narawita et al., 2017) | Tokenization<br>POS<br>Chunking<br>Splitting | Weka vote algorithm<br>Logistics and SMO | English text | Usecase diagram<br>Class diagram | - | Visual studio<br>Weka | Unstructured | Accuracy<br>Applicability |